\documentclass{article}

\usepackage{arxiv}

\usepackage[utf8]{inputenc} 
\usepackage[T1]{fontenc}    
\usepackage{hyperref}       
\usepackage{url}            
\usepackage{booktabs}       
\usepackage{amsfonts}       
\usepackage{nicefrac}       
\usepackage{microtype}      
\usepackage{lipsum}		    
\usepackage{graphicx}
\usepackage{natbib}
\usepackage{doi}

\usepackage{amsmath,amssymb,amsfonts,mathtools, comment}
\usepackage{algorithmic}
\usepackage{graphicx}
\graphicspath{{./images/}}
\usepackage{textcomp}
\usepackage{wrapfig}
\usepackage{subcaption}
\usepackage{bm}
\usepackage{multirow}
\usepackage{soul}
\usepackage{array}
\newcolumntype{M}[1]{>{\centering\arraybackslash}p{#1}}
\newcommand\norm[1]{\left\lVert#1\right\rVert}
\DeclareMathOperator*{\argmin}{argmin}

\title{Dynamic Semantic VSLAM with Known and Unknown Objects}

\author{Sanghyoup Gu \\
	Department of Electrical and Computer Engineering\\
	Iowa State University\\
	Ames, IA 50014 \\
	\texttt{rndk9004@iastate.edu} \\
	\And
	Ratnesh Kumar \\
	Department of Electrical and Computer Engineering\\
	Iowa State University\\
	Ames, IA 50014 \\
	\texttt{rkumar@iastate.edu} \\
}

\renewcommand{\shorttitle}{\textit{arXiv} Template}

\hypersetup{
pdftitle={A template for the arxiv style},
pdfsubject={q-bio.NC, q-bio.QM},
pdfauthor={David S.~Hippocampus, Elias D.~Striatum},
pdfkeywords={First keyword, Second keyword, More},
}

\begin{document}
\maketitle

\begin{abstract}
Traditional Visual Simultaneous Localization and Mapping (VSLAM) systems assume a static environment, which makes them ineffective in highly dynamic settings. To overcome this, many approaches integrate semantic information from deep learning models to identify dynamic regions within images. However, these methods face a significant limitation as a supervised model cannot recognize objects not included in the training datasets. This paper introduces a novel feature-based Semantic VSLAM capable of detecting dynamic features in the presence of both known and unknown objects. By employing an unsupervised segmentation network, we achieve unlabeled segmentation, and next utilize an objector detector to identify any of the known classes among those. We then pair this with the computed high-gradient optical-flow information to next identify the static versus dynamic segmentations for both known and unknown object classes. A consistency check module is also introduced for further refinement and final classification into static versus dynamic features. Evaluations using public datasets demonstrate that our method offers superior performance than traditional VSLAM when unknown objects are present in the images while still matching the performance of the leading semantic VSLAM techniques when the images contain only the known objects. 
\end{abstract}


\section{Introduction}
\label{sec:introduction}
Simultaneous Localization and Mapping (SLAM) aims to estimate the mobile camera's pose/location while simultaneously reconstructing a map of its unknown environment in the absence of any inertial measurement unit (IMU) \cite{VSLAM2022}. Owing to low cost, lightweight, and compact size of cameras, the Visual SLAM (VSLAM) has drawn much attention, making it suitable for various applications such as autonomous vehicles\cite{SlamInautonomous}, augmented reality\cite{SlamInAR}, search and rescue\cite{SlamSAR_2020}, and precision agriculture\cite{SlamInTheField}.

The traditional VSLAM systems assume a static environment, where all elements in the scene are either static or moving slowly. However, this limits VSLAM's applicability in real-world scenarios possessing various dynamic components\cite{Tourani_2022}. While most VSLAM systems are implemented with robust estimation techniques like RANSAC\cite{RANSAC_1981} to handle a small number of outliers, their localization accuracy significantly drops as the proportion of dynamic regions in an image increases. 

To overcome the static world requirement of the traditional VSLAM, dynamic VSLAM methods develop algorithms to distinguish dynamic areas in image sequences using prior information or constraints. In particular, the recent advancements in deep learning and computer vision have facilitated the use of semantic information to identify dynamic objects, leading to enhanced localization performance and improved situational awareness for robots\cite{Chen2022}. In this paper, semantic information refers to detection methods that utilize object class, shape, location, etc. information in an image. Consequently, newer dynamic VSLAM systems integrate semantic information for navigation. Such semantic VSLAM methods typically follow a two-stage approach for classifying static vs. dynamic features: Initially, the features are categorized in static vs. dynamic for the known object classes, where the features on movable objects are categorized as quasi-dynamic, and the other features are categorized as quasi-static. Subsequently, a rough camera pose is estimated from the quasi-static features, and next certain geometric constraints are applied to validate the dynamic nature of quasi-dynamic features. 
These semantic VSLAM methods have a critical limitation in that their performance heavily depends on the labeled training dataset, and they fail to detect objects that are not included in the training dataset. For instance, the well-known COCO\cite{COCO} and PASCAL VOC\cite{PASCALVOC} detection datasets include only 80 and 20 classes, respectively. Semantic VSLAM utilizing models trained on these datasets can only recognize the predefined classes, whereas the real world comprises many other objects.

To address this issue, in this work, we propose a dynamic VSLAM framework capable of handling static and dynamic objects of both known and unknown kinds. The proposed system is built on ORB-SLAM2, a reliable feature-based VSLAM system, augmented with a new classification module to distinguish dynamic features. This classification module includes segmentation plus optical flow threads. We employ an unsupervised segmentation model, yielding segmentations for all objects, known or unknown. Additionally, we also compute the optical flow. We then label dynamic segmentation based on either known movable objects as determined from an integrated object detector, or from the computed high optical flow gradients. The figure in the abstract illustrates the proposed method's result: Detecting dynamic features within known objects (such as people) as well as unknown objects (such as boxes). The contributions of this paper are as follows:

\begin{itemize}
    \item A modified unsupervised segementer, Fast-SAM \cite{FastSAM} is introduced that provides segmentation for all objects, known as well as unknown, allowing the system to leverage image information without being restricted to known object classes. This is further paired with an object detection head to identify any of the known objects, and thus yielding labeled segmentation for those objects that are identified as known.
    \item SDO (Semantic, detection, and Optical-flow based) classification algorithm based on the above partially labeled segmentation output coupled with the computed optical flow gradients is proposed, capable of detecting dynamic areas even for objects absent from the training dataset.
    \item An additional consistency check module is introduced that estimates the scene flow and uses that to refine the static versus dynamic classification, yielding a final robust result.
    \item The proposed method's performance is validated using public datasets containing known and unknown dynamic objects. Our method offers superior performance than traditional VSLAM when unknown objects are present in the images, while still matching the performance of the leading semantic VSLAM techniques when the images contain only the known objects. 
\end{itemize}

The remainder of this paper is organized as follows. Section \uppercase\expandafter{\romannumeral2} reviews the related works. Section \uppercase\expandafter{\romannumeral3} presents our proposed semantic VSLAM method for identifying static vs. dynamic features from both known and unknown objects. Section \uppercase\expandafter{\romannumeral4} discusses evaluation results from TUM RGB-D\cite{TUMRGBD} and OMD (Oxford Multi-motion Dataset)\cite{OMD}. Section \uppercase\expandafter{\romannumeral5} provides a conclusion.

\section{Related Works}
\begin{figure} [h!]
\centerline{\includegraphics[width=0.6\textwidth]{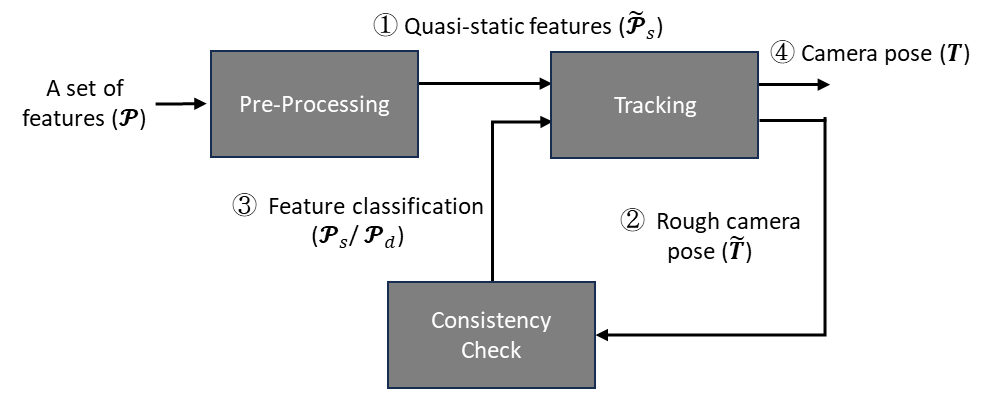}}
\caption{Two stage approach}
\label{Two stage approach}
\end{figure}

Performing a concurrent identification of mobile camera pose and static vs. dynamic features when there is no on-board IMU (inertial measurement unit) presents a cyclical problem: Estimating the mobile camera pose is essential for identifying the static vs. dynamic features, but the mobile camera pose estimation process relies on the use of static features. Dynamic VSLAM systems typically tackle this issue with a two-stage method: Initially, a set of features is identified, and the subset of quasi-static features---assumed to be mostly static---are selected using prior information like semantic class or geometric constraints in the ``pre-processing module" (see Fig. \ref{Two stage approach}). ``The tracking module" (of Fig. \ref{Two stage approach}) uses these quasi-static features to estimate an initial rough camera pose. Subsequently, the feature classification is refined through certain means (``consistency check", also see Fig. \ref{Two stage approach}) that vary among approaches, leading to a final camera pose estimation using the updated static features. 

Dynamic VSLAM systems are categorized into non-Semantic vs. Semantic methods. Non-semantic methods utilize geometric constraints or optical flow to determine dynamic versus static features. It is assumed that geometric constraints from camera motion apply to a larger number of static features while a smaller number of dynamic features exhibit different geometric patterns. Also, optical flow indicates the pixel movement between frames, offering clues for moving areas, subject to overlapping effects of camera vs. object motions. 

\begin{itemize}
    \item 
    \cite{SUN2017110} initially estimated the pose using all features from two images, detecting dynamic objects through ego-motion compensated image differencing, where non-zero pixels in the difference image were taken to indicate motion, and further refined the estimate by tracking using a particle filter. 
    \item 
    \cite{DGFlow-SLAM} divided an image into small grids, categorizing them as dynamic or static based on scene flow computed from optical flow and previous camera motion, with iterative pose refinement, excluding dynamic features.  
    \item 
    \cite{ObjectSegmentationFromOpticalFlow} identified motion boundaries from optical flow gradients, refining initial motion boundaries using appearance models across the video. 
    \item 
    \cite{MovingModel} minimized optical flow error for camera pose estimation, deriving posterior probability models from flow angle likelihood conditional on optical flow and camera pose. 
\end{itemize}

\begin{figure*} [t!]    
\centerline{\includegraphics[width=0.95\textwidth]{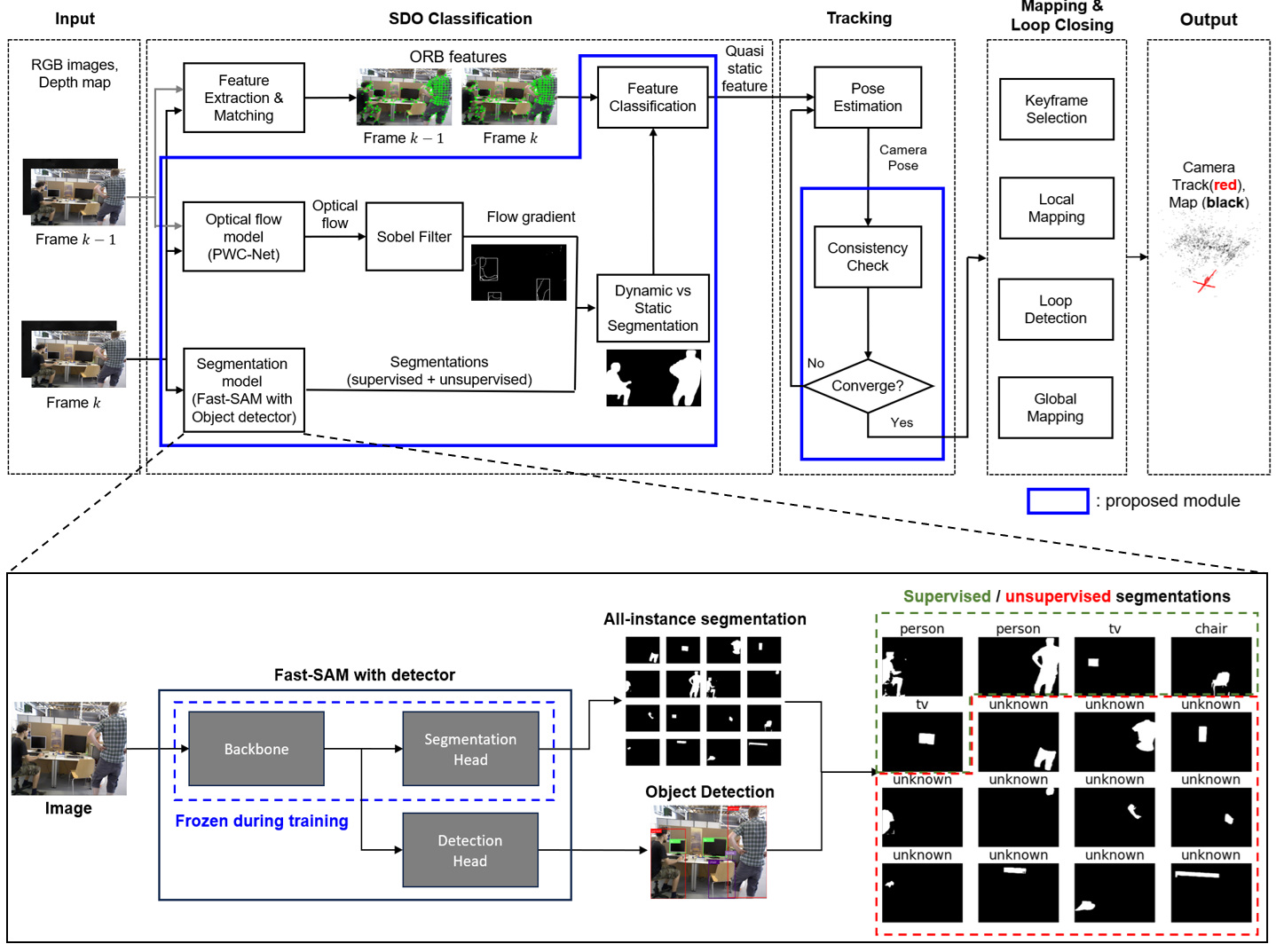}}
\caption{Framework and segmentation model}
\label{fig: Framework}
\end{figure*}

The non-Semantic approaches do not use any semantic prior (i.e., information regarding object class, shape, location, etc.), and mostly predate the development of semantic approaches. The non-semantic approach has significantly decreased localization performance as the portion of dynamic regions in a scene increases. Also, optical flow errors increase with depth, further complicating object motion detection. In addition, since the optical flows result from both camera and object movements, excluding camera motion flow and detecting object motions have circular dependency due to the association of object flow with camera motion, depth, and pixel position. 

The semantic approach uses deep-learning models like YOLO\cite{YOLO3}, Mask R-CNN\cite{MaskRCNN}, SegNet\cite{SegNet}, and SSD\cite{SSD} for semantic segmentation to then identify dynamic features from known movable objects (e.g., people, cars, bicycles) and exclude those during the rough camera pose estimation. Then, the pose is refined using additional geometric or optical flow constraints.

\begin{itemize}
    \item 
    \cite{DynaSLAM} utilized Mask R-CNN to identify movable objects. The depth differences between the current frame and projected frames from nearby keyframes were computed to estimate dynamic features. After excluding the dynamic features, it inpainted the excluded area in the map. 
    \item 
    \cite{DS-SLAM} and 
    \cite{SOFSLAM} used SegNet to detect movable objects and refined pose estimation using an epipolar constraint, considering features far from the epipolar line as dynamic.
    \item 
    \cite{DynamicSLAMwithYOLO5} applied YOLO-v5s for movable object detection and used optical flow to distinguish quasi-dynamic features. 
    \item 
    \cite{VDO-SLAM} used Mask R-CNN and PWC-Net\cite{PWCNet} to obtain semantic information and dense optical flow, respectively.     Then, features inside the movable objects were classified using scene flow computed from optical flow and together with the estimated rough camera pose. 
\end{itemize}

While the advances in deep learning have enhanced object detection accuracy in the presence of moving objects, unknown objects that cannot be recognized by supervised models end up compromising the overall VSLAM system's efficacy for their reliance then on geometric constraints for static vs. dynamic feature classification. 
These challenges then motivate our proposed approach presented next.

\section{Proposed Approach for Static vs. Dynamic Features}
Given the above-listed existing approaches to dynamic VSLAM and their respective limitations, our proposed system introduces a new classification module to distinguish dynamic features on top of the existing ORB-SLAM2 system, a feature-based VSLAM that is known to work well for static environments. The other modules, such as mapping and loop closing, remain consistent with ORB-SLAM2. An overview of the proposed system is presented in Fig. \ref{fig: Framework}, in which the newly introduced modules are marked with a blue outline. 

\begin{itemize}
    \item Initially, a current RGB image is fed into an unsupervised segmentation model to obtain the image segmentations, where we have introduced an object detection head to the unsupervised segmentation module, thereby further yielding both known and unknown objects' segmentations (see the left bottom module inside the "SDO classification" module in Fig. \ref{fig: Framework} and its blowup as well as the details in Section \uppercase\expandafter{\romannumeral3}-A).
    \item Simultaneously, current and previous images are processed by an optical flow model to gather pixel-wise optical flow information, and next, the optical flow gradient is computed using the Sobel filter, where the object motion boundaries are identified based on the high optical flow gradient (see the left middle module inside the "SDO classification" module in Fig. \ref{fig: Framework} and the details in Section \uppercase\expandafter{\romannumeral3}-B)
    \item By combining optical flow gradient-based motion boundaries with the segmentations, an initial set of static vs. dynamic segmentations is estimated (see the right bottom module inside the ``Classification" module in Fig. \ref{fig: Framework}, with details in Section \uppercase\expandafter{\romannumeral3}-C).
    \item Next, using the estimated static features, a rough camera pose is estimated. (see the top module inside the ``Tracking" module in Fig. \ref{fig: Framework}, with details in Section \uppercase\expandafter{\romannumeral3}-D)
    \item Following an initial rough camera pose estimation from the initial estimate of static features, the feature classification is refined through the proposed iterative consistency check module that evaluates the scene flow for refinement. Finally, the camera pose is re-estimated using the refined set of static features (see the 2nd module from the top inside the ``Tracking" module in Fig. \ref{fig: Framework}, with details in Section \uppercase\expandafter{\romannumeral3}-E).
\end{itemize}

\subsection{Segmentation}
For the sake of real-time application of VSLAM, Fast-SAM\cite{FastSAM}, comprising a backbone and a segmentation head, is employed for unsupervised segmentation due to its 50x faster inference time compared to SAM (Segment Anything Model)\cite{SAM} by utilizing a CNN backbone from YOLOv8 (instead of the vision transformer backbone from SAM). Unlike the supervised models that provide segmentations for known objects, Fast-SAM offers all potential segmentations without any class labels.
The RGB image is transformed into a 3D feature map by the CNN backbone, which is then processed by the segmentation head to produce segmentations, $\textit{S}\in\{0,1\}^{w \times h \times n_{s}}$ where $w$ and $h$ denote the image width and height, and $n_{s}$ represents the number of segmentations. To achieve both supervised object detection and unsupervised segmentation, we integrate the YOLOv8 detection head with the Fast-SAM backbone, as depicted in Fig. \ref{fig: Framework}. The detection head is trained using the COCO detection dataset while other Fast-SAM layers remain frozen to maintain the segmentation performance of the original Fast-SAM model. The training configuration, including the loss function and optimizer, mirrors that of the YOLOv8 model. The detection head identifies the object class and bounding box for only the known objects, while the segmentation head provides pixel-wise segmentation for known and unknown objects. Segmentations with the highest Intersection over Union (IoU) with known object boxes are assigned the corresponding object classes and bounding boxes. The output of this step thereby includes both known and unknown objects segmentations.

\subsection{Optical flow gradient}
Optical flow describes the two-dimensional displacement of pixels between two images, offering clues for motion. Since the optical flow arises from both camera and object motions, one need to somehow decouple the two effects to the best possible. Our key observation is that the {\em optical flow gradients} can provide clearer clues for the object motion than the bare optical flows themselves. This is based on the fact that neighboring pixels from static landmarks move similarly, solely due to camera motion, and so there exist boundaries of optical flow among static vs. dynamic features. When two consecutive images are provided, the optical flow is estimated using PWC-Net\cite{PWCNet}, and subsequently, the Sobel filter is applied to derive the corresponding optical flow gradient. 

\begin{figure} [h!]
\centerline{\includegraphics[width=0.7\textwidth]{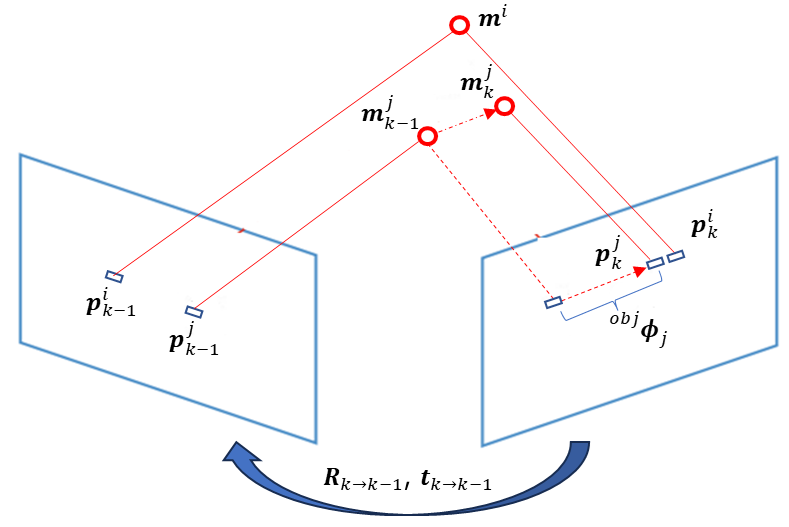}}
\caption{Optical flow gradient}
\label{Optical flow gradient}
\end{figure} 
Fig. \ref{Optical flow gradient} demonstrates the formation of optical flow gradient for static and dynamic features. In Fig. \ref{Optical flow gradient}, the right superscript and subscript denote index and time, respectively: $\boldsymbol{p}^{i}_{k}$ is $i$-th feature at time $k$. $\boldsymbol{R}_{k \rightarrow k-1}$, and $\boldsymbol{t}_{k \rightarrow k-1}$ indicate camera rotation and translation from time $k$ to $k-1$, respectively. A landmark $\boldsymbol{m} \in \mathbb{R}^3$ is projected onto a pixel $\boldsymbol{p} \in \mathbb{R}^2$ in the image plane via the camera projection equation: 
\begin{align}  
\left[\begin{array}{c}
\boldsymbol{p}\\1\end{array}\right] = \pi(\boldsymbol{m})=\frac{1}{d}\boldsymbol{K}\boldsymbol{m}\Leftrightarrow \boldsymbol{m}=dK^{-1}\left[\begin{array}{c}
\boldsymbol{p}\\1\end{array}\right], \label{eq1}
\end{align}
where $\boldsymbol{K} \in \mathbb{R}^{3 \times 3}$ is camera intrinsic, $d \in \mathbb{R}$ is the landmark depth, and the landmark is represented in the camera body frame.

For a static landmark, their positions at two frames are associated with the ensuing camera rotation and translation:
\begin{align}  \boldsymbol{m}_{k-1} = \boldsymbol{R}_{k \rightarrow k-1}\boldsymbol{m}_{k}+\boldsymbol{t}_{k \rightarrow k-1}. \label{eq2}
\end{align}
Then the optical flow for a pixel due to only the camera motion is computed as follows, where for simplified notation, the subscript of rotation and translation is omitted ($\boldsymbol{R}\equiv\boldsymbol{R}_{k \rightarrow k-1},\boldsymbol{t}\equiv\boldsymbol{t}_{k \rightarrow k-1}$):
\begin{align} 
\left[\begin{array}{c}
\boldsymbol{\prescript{cam}{}{\boldsymbol{\phi}_{k}}}\\1\end{array}\right]
 & = \textstyle \left[\begin{array}{c}
\boldsymbol{p_k}\\1\end{array}\right] - \left[\begin{array}{c}
\boldsymbol{p_{k-1}}\\1\end{array}\right] \nonumber\\ 
& = \frac{1}{d_{k}}\boldsymbol{K}\boldsymbol{m}_{k} - \frac{1}{d_{k-1}}\boldsymbol{K}\boldsymbol{m}_{k-1} \nonumber\\
& = \frac{1}{d_{k}}\boldsymbol{K}\boldsymbol{m}_{k} - \frac{1}{d_{k-1}}\boldsymbol{K}\left( \boldsymbol{R}\boldsymbol{m}_{k}+\boldsymbol{t} \right) \nonumber\\
& = \boldsymbol{K} \textstyle \left(\boldsymbol{I}-\frac{d_{k}}{d_{k-1}}\boldsymbol{R}\right)\frac{\boldsymbol{m}_{k}}{d_{k}} - \frac{1}{d_{k-1}}\boldsymbol{K}\boldsymbol{t} \nonumber\\
& = \boldsymbol{K} \textstyle \left(\boldsymbol{I}-\frac{d_{k}}{d_{k-1}}\boldsymbol{R}\right)\boldsymbol{K}^{-1}\left[\begin{array}{c}
\boldsymbol{p_k}\\1\end{array}\right] - \frac{1}{d_{k-1}}\boldsymbol{K}\boldsymbol{t} \nonumber\\
& \approx \boldsymbol{K}\left(\boldsymbol{I}-\boldsymbol{R}\right)\boldsymbol{K}^{-1}\left[\begin{array}{c}
\boldsymbol{p_k}\\1\end{array}\right] - \frac{1}{d_{k-1}}\boldsymbol{K}\boldsymbol{t}. 
\label{eq3}
\end{align}

Note in the last approximation, the effect of camera motion on depth change between the adjacent time steps is taken to be negligible ($d_{k-1} \approx d_{k}$).

Next, to compare the optical flow gradient between the dynamic object and the static background, consider two neighboring features $i$ and $j$ projected from the static vs. dynamic landmarks, respectively. A static feature's optical flow is  generated from only the camera flow, whereas a dynamic feature's optical flow is generated from object as well as camera motions:
\begin{align}  \boldsymbol{\phi}^{i}_{k} = \prescript{cam}{}{\boldsymbol{\phi}^{i}_{k}};\quad \boldsymbol{\phi}^{j}_{k} = \prescript{obj}{}{\boldsymbol{\phi}^{j}_{k}} + \prescript{cam}{}{\boldsymbol{\phi}^{j}_{k}}. \label{eq4} 
\end{align} 
The {\em optical flow gradient} between two adjacent features $i$ and $j$ is then computed applying \eqref{eq3}  and \eqref{eq4}, in which the first term of camera motion gets canceled for adjacent features ($\boldsymbol{p}^{i}_{k} \approx \boldsymbol{p}^{j}_{k}$):
\begin{align} \boldsymbol{\phi}_{k}^{j}-\boldsymbol{\phi}_{k}^{i} \approx  \prescript{obj}{}{\boldsymbol{\phi}^{j}_{k}}-\textstyle\left(\frac{1}{d_{k-1}^{j}}-\frac{1}{d_{k-1}^{i}}\right)\boldsymbol{K}\boldsymbol{t}. \label{eq5}
\end{align} 
(It should be noted here that since $i$ and $j$ are adjacent features, the two indices differ by 1, and so the optical flow gradient at $i$ vs. $j$ is the same as the optical flow difference at those two features, as computed in \eqref{eq5}.) 

From the aforementioned derivation, we can see that the optical flow gradient among adjacent pixels is independent of pixel position. Further, for two objects of similar depth (so the second term in \eqref{eq5} is negligible), the optical flow gradient is dominated by the optical flow of the dynamic object. Hence identifying a high optical flow gradient serves to indicate motion boundaries, allowing for the potential aberration introduced from any large depth variations among neighboring pixels of moving vs. static features (so the second term in \eqref{eq5} can no longer be neglected, affecting the overall optical flow gradient). To reduce the corresponding aberration in identifying static versus dynamic features, 
we develop an iterative approach for estimating the camera pose by first excluding the initially identified dynamic features and next relying solely on the remaining static features we compute a rough initial estimate of the camera pose. Later, we perform a consistency check by computing scene flow based on the initial camera pose estimate and discard the outliers (those having large non-zero scene flow) to redo the camera pose estimation. The two steps are iterated until convergence.

\subsection{SDO (Semantic, Detection, and Optical-flow based) Classification}
\begin{figure} [h!]
\centerline{\includegraphics[width=0.7\textwidth]{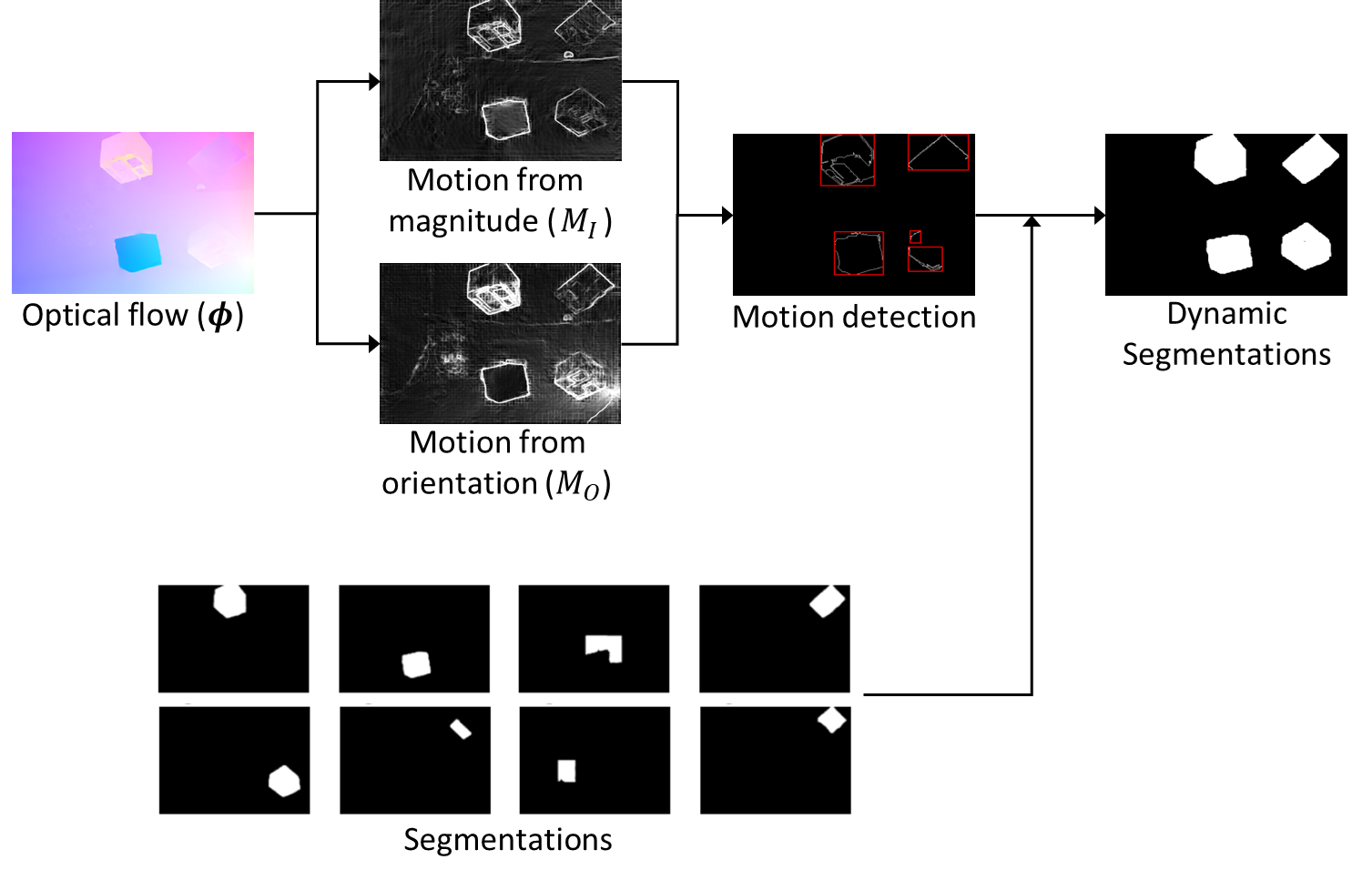}}
\caption{Dynamic segmentations selection}
\label{Dynamic segmentations selection}
\end{figure}

We assign dynamic segmentation labels based on either known movable object segmentation or high optical flow gradients; the rest of the segmentations are labeled static. The modified Fast-SAM model detects 80 classes from the COCO dataset. Animals or vehicles, which can move, are categorized as movable objects and excluded from camera pose estimation. Segmentations with high optical flow gradients are also excluded. Motion boundaries, derived from optical flow gradients, are computed as in \cite{ObjectSegmentationFromOpticalFlow}: 
\begin{align} & M^{i}_{I} = 1 - exp\left(-k_{I}\|\boldsymbol{\phi}^{i}\|\right) \\ 
              & M^{i}_{O} = 1 - exp\left(-k_{O}\max_{j\in N}|\delta\theta_{i,j}| \right), 
\end{align} 
where $M^{i}_{I},  M^{i}_{O}\in[0,1]$ are the motion boundaries from gradient magnitude and orientation difference respectively, at pixel $i$; $k_{I}$ and $k_{O}$ are parameters to control the steepness, $N$ is the set of neighboring pixels surrounding pixel $i$, and $\delta\theta_{i,j}$ is orientation difference between optical flow at pixel $i$ and $j$. These two boundaries are combined as follows: 
\begin{align} M^{i} = \bigg \lbrace\begin{array}{ll} M^{i}_{I} & M^{i}_{I}> 0.8 \\                                
                                                     M^{i}_{I}M^{i}_{O} & otherwise. \end{array}   
\end{align} 
The combined motion boundaries are transformed into a binary map using a threshold of 0.5. Dynamic areas are then identified by detecting contours within this binary motion map. 
Segmentations having a maximum overlapped area with the detected motion boundaries are selected as dynamic segmentations. Fig. \ref{Dynamic segmentations selection} shows the dynamic segmentation selection from high optical flow gradients. In the figure, motion detection from a high optical flow gradient is not precisely aligned with boxes because of disconnected contours from optical flow gradients. However, we are able to obtain more precise dynamic segmentation by labeling dynamic segmentation from the detected motion boundaries.

\begin{figure*}[ht!]
    \centering
    \begin{subfigure}{0.48\linewidth}
        \centering
        \includegraphics[width=8.2cm]{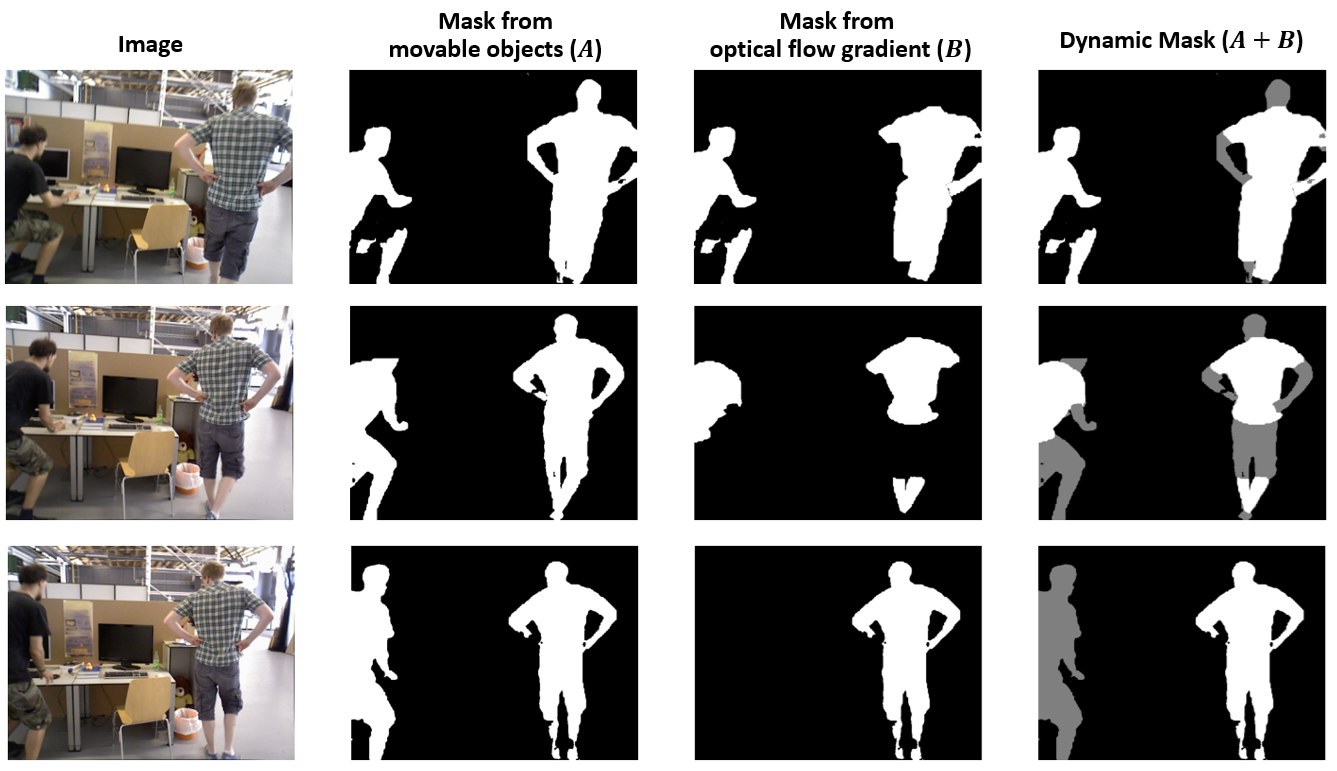}
        \caption{Dynamic Mask for TUM}
        \label{fig: Dynamic Mask for TUM}
    \end{subfigure}
    \begin{subfigure}{0.48\linewidth}
        \centering   
        \includegraphics[width=8.2cm]{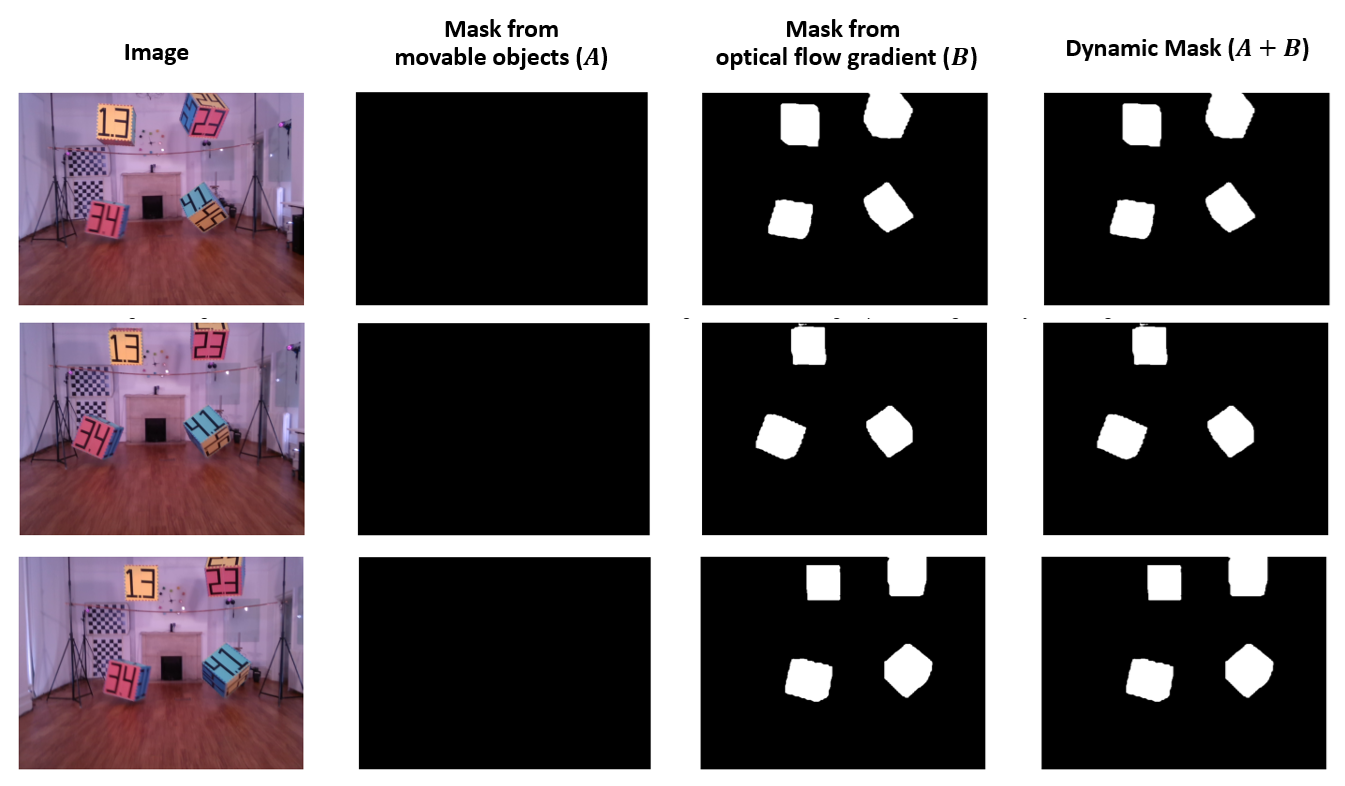}
        \caption{Dynamic Mask for OMD}
        \label{fig: Dynamic Mask for OMD}
    \end{subfigure}
\caption{Results of dynamic mask selection}
\label{Dynamic Mask}
\end{figure*}

Dynamic segmentations from known movable objects versus from high optical flow gradients complement each other: Consider for example Fig. \ref{Dynamic Mask} that shows the detection result for two popular datasets. In the TUM dataset, a person is segmented partially because only part of the person moves. But by adding known movable object segmentation, complete segmentation for the person gets to be found. In the case of the OMD dataset, boxes, that are unknown objects, are not detected at all from movable object segmentation, but they are detected from the optical flow gradient. 

After obtaining static vs. dynamic segmentations from movable objects and high optical flow gradients, dynamic segmentation features are classified as quasi-dynamic, while the other features are classified as quasi-static.

\subsection{Camera Pose Estimation}
An initial rough camera pose is estimated from the quasi-static features. The initial camera pose at time $k$ (rotation $\hat{\boldsymbol{R}}_k^-$ and translation $\hat{\boldsymbol{t}}_k^-$) is estimated by minimizing the reprojection error between the estimated stationary landmarks and corresponding features in the image:

\begin{align}  \{\hat{\boldsymbol{R}}_k^-, \hat{\boldsymbol{t}}_k^-\}\!=\! \argmin_{\boldsymbol{R}, \boldsymbol{t}}\!\!\!\! \sum_{i \in \hat P_{s,k}^-}\!\!\!\rho\biggl(\norm{\left[\!\!\begin{array}{c}
\boldsymbol{p_k^i}\\1\end{array}\!\!\right]-\pi(\boldsymbol{R}
\prescript{W}{}{\boldsymbol{m}^{i}_{k-1}}
+\boldsymbol{t})}^{2}_{\Sigma}\!\biggl),\label{eq9}
\end{align}
where an estimated stationary landmark $
\prescript{W}{}{\boldsymbol{m}^{i}_{k-1}}
\in \mathbb{R}^{3}$ is in the world frame and is computed from previous camera pose and its position in the body frame: $
\prescript{W}{}{\boldsymbol{m}^{i}_{k-1}}=\hat{\boldsymbol{R}}_{k-1}\boldsymbol{m}^{i}_{k-1}+\hat{\boldsymbol{t}}_{k-1}$, in which $\boldsymbol{m}^{i}_{k-1}=d^i_{k-1}K^{-1}\left[\!\!\begin{array}{c} \boldsymbol{p^i_{k-1}}\\1\end{array}\!\!\right]$ is found from its matched feature $\boldsymbol{p}^{i}_{k-1}$ of the previous frame using \eqref{eq1}, $\rho(\cdot)$ is the robust Huber function, $\Sigma$ is the covariance matrix associated with the reprojection error, and $\hat P_{s,k}^-$ is the initially estimated index set of static features at time $k$.

\subsection{Consistency check and classification/pose Refinement}
The initial estimate of the static features may include certain dynamic segmentations from unknown object classes as they may possess low optical flow gradients for being reduced from higher depth gap effects. [Dually, the initial estimation may also exclude certain static features, again because of the depth gap effects leading to high optical flow gradients. But those do not used in the camera pose estimation and so are benign.]
To refine feature classification (and ensuing pose estimation), we introduce a consistency check step involving computing the scene flow, representing the three-dimensional displacement of landmarks between consecutive frames:
\begin{align} f^{i}_{k} & = \prescript{W}{}{\boldsymbol{m}^{i}_{k}} - \prescript{W}{}{\boldsymbol{m}^{i}_{k-1}} \nonumber \\
                         & \approx \hat{\boldsymbol{T}}_{k}^- \left[\begin{array}{c} 
                         {\boldsymbol{m}^{i}_{k}}\\1\end{array}\right] - \hat{\boldsymbol{T}}_{k-1}\left[\begin{array}{c} 
                         {\boldsymbol{m}^{i}_{k-1}}\\1\end{array}\right] ,k\geq 1
\end{align} 
Here $\hat{\boldsymbol{T}}_{k}^-:=\left[\begin{array}{ll}\hat{\boldsymbol{R}}_k^-&\hat {\boldsymbol{t}}_k^-\\0_{1\times 3}&1\end{array}\right]$ is the rough camera pose estimated from rough quasi-static features $\hat P_{s,k}^-$ at time step $k$ as in \eqref{eq9}, and $\hat{\boldsymbol{T}}_{k-1}$ is the refined camera pose estimated from refined quasi-static features $\hat P_{s,k-1}$ at time $k-1$ with $\hat{\boldsymbol{T}}_0:= I$. 

A scene flow directly indicates whether or not a landmark is static since the static landmarks will exhibit near-zero scene flow. Accordingly, the features whose scene flows exceed a threshold are classified as dynamic. Accordingly, the segmentations are re-evaluated: A segmentation is deemed dynamic if it includes more than thirty percent dynamic features, and otherwise, it is classified as static segmentation. (Additionally, the segmentations for the movable objects are considered as dynamic regardless of the number of dynamic features.)

Next, using the newly refined classes of static versus dynamic features, dynamic segmentations are excluded to obtain the refined estimated index set of static features $\hat P_{s,k}$ for the sake of the refined camera pose estimation $\hat{\boldsymbol{T}}_{k}$. For this, the same estimation equation of \eqref{eq9} is used but by replacing $\hat P_{s,k}^-$ with $\hat P_{s,k}$ (the refined set of static features).  These steps of static feature refinement and camera pose estimation are iterated until the static vs. dynamic classification converges. 

\begin{figure} [h!]
\centerline{\includegraphics[width=0.7\textwidth]{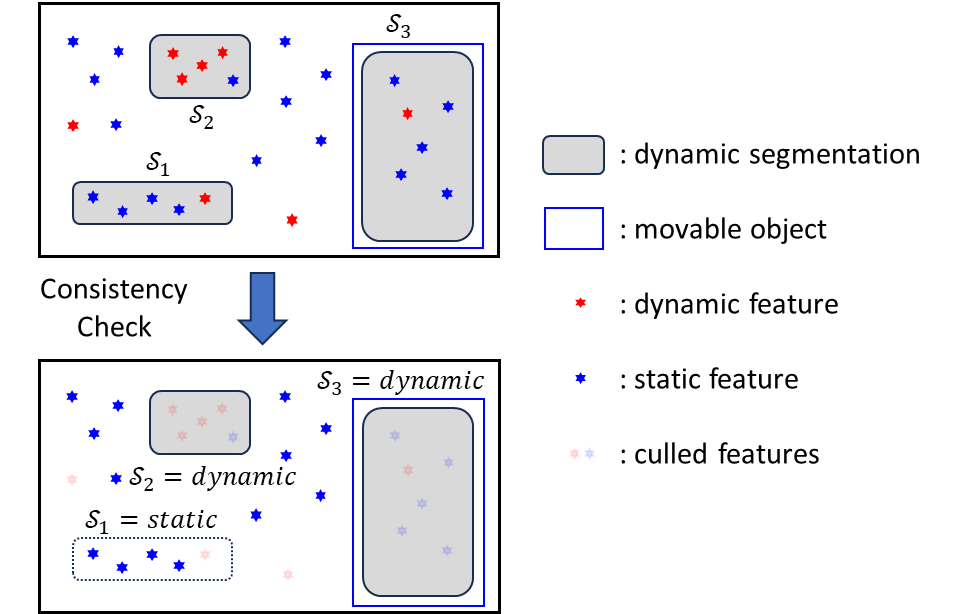}}
\caption{Consistency check and Segmentation Refinement}
\label{fig: refine}
\end{figure}

Fig. \ref{fig: refine} shows the result of the iterative consistency check. Features in movable objects or dynamic segmentations are excluded for pose estimation. In addition, if features outside movable objects and dynamic segmentations have scene flow greater than the threshold, these features are also excluded. By culling these features, the camera pose is then estimated using the new set of static features.

\section{Evaluation}

The proposed method has been evaluated using the TUM and OMD public tracking datasets. The TUM RGB-D dataset, known for its significant dynamic factors in indoor environments, predominantly includes dynamic elements whose classes are predefined in the training dataset. This dataset is commonly used to validate dynamic VSLAM methods, making it suitable for comparing the performance of the proposed method against the existing dynamic VSLAM techniques. In contrast, the OMD dataset provides indoor scenes with dynamic features and unknown objects.  The OMD dataset is used to assess performance in the presence of unknown dynamic objects: It features swinging boxes, which are not included in the training dataset and hence belong to the unknown class, and the supervised models like Mask R-CNN and SegNet fail to detect those. 

Our proposed method's performance is compared to conventional VSLAM (ORB-SLAM2), non-semantic dynamic VSLAM, and semantic dynamic VSLAMs (Dyna-SLAM and DS-SLAM) on the TUM dataset with known dynamic objects. 
Thereby the impact of our proposed feature classification in the dynamic environments with known objects is validated against the existing ones on the TUM dataset. Further, our method is compared to conventional VSLAM for the OMD dataset to compare feature classification performance when there also exist unknown moving objects in images.        

To compare the performance of multiple VSLAM systems, we use ATE (Absolute Trajectory Error) and RTE (Relative Trajectory Error)\cite{evaluation} for the camera motion as the evaluation metrics. ATE measures the absolute cumulative error between the ground truth camera trajectory $\boldsymbol{T}_k$ versus the estimated camera trajectory $\hat{\boldsymbol{T}}_k$, comparing the closeness of the estimated trajectory with the ground truth one. To begin, the transformation $\boldsymbol{S} \in SE(3)$ from the estimated trajectory onto the ground truth trajectory is obtained using the least-squares solution. Given this transformation, the absolute error at time step $k$ is computed as:
\begin{align}  \boldsymbol{F}_{k} = \boldsymbol{T}_{k}^{-1}\boldsymbol{S}\boldsymbol{\hat{T}}_{k},     
\end{align} 
and the absolute error of all time indices is computed by the RMSE (root mean squared error):
\begin{align} RMSE(\boldsymbol{F}_{1:n}) & = \biggl( \frac{1}{n} \sum_{k=1}^{n} \norm{trans(\boldsymbol{F}_k)}^2 \biggl) ^{1/2}, 
\end{align} 
where $n$ is the number of trajectory points, and $trans(\cdot) \in \mathbb{R}^{3}$ computes the translation part of a pose matrix.

Since an initial camera position error in estimation may affect to exaggerate future position errors, RTE is also used as a comparison metric, complementing ATE. RTE at any time step $k$ compares relative pose change within a fixed number of time steps $\Delta$:
\begin{align}  \boldsymbol{E}_{k} = \biggl(\hat{\boldsymbol{T}}_{k}^{-1}\hat{\boldsymbol{T}}_{k+\Delta}\biggl)^{-1} \biggl(\boldsymbol{T}_{k}^{-1}\boldsymbol{T}_{k+\Delta} \biggl),
\end{align} 
and similar to the ATE, the RTE of all time indices is computed by way of RMSE:
\begin{align} RMSE(\boldsymbol{E}_{1:n}, \Delta) & =\left[\!\! \begin{array}{ll} \bigg( \frac{1}{n-\Delta} \sum_{k=1}^{n-\Delta} \norm{trans(\boldsymbol{E}_k)}^2 \bigg) ^{1/2} \\ 
\bigg( \frac{1}{n-\Delta} \sum_{k=1}^{n-\Delta} |rot(\boldsymbol{E}_k)|^2 \bigg) ^{1/2}   \end{array}\!\! \right],
\end{align}
where $rot(\cdot) \in SO(3)$ computes the rotation part of a pose matrix. The magnitude of the rotation matrix is measured by the corresponding rotation angle $|rot(\boldsymbol{E}_i)| = arccos(\frac{tr(\boldsymbol{E}_i)-1}{2})$. 

Each trajectory sequence is run five times to account for non-deterministic components in the VSLAM methods, and the median value is used for performance comparison as summarized in Tables~\ref{Table 1} and \ref{Table 2}.

\begin{table*}[ht!]
\centering
\caption{ATE for TUM dataset}
\label{Table 1}
  \centering
  \resizebox{\linewidth}{!}{ 
\renewcommand{\arraystretch}{2.0}
\begin{tabular}{M{0.2\linewidth} | M{0.16\linewidth} M{0.16\linewidth} M{0.16\linewidth} M{0.16\linewidth} M{0.16\linewidth}}
\hline
\hline
\multirow{2}{*}{\textbf{Sequence}}  & \textbf{ORB-SLAM2 \cite{ORB-SLAM2}} & \textbf{SLAM with motion removal \cite{SUN2017110}} &  \textbf{Dyna-SLAM \cite{DynaSLAM}} & \textbf{DS-SLAM \cite{DS-SLAM}} & \textbf{Our method}  \\ \cline{2-6}
                               & RMSE (m) & RMSE (m) &  RMSE (m) & RMSE (m) & RMSE (m) \\ 
\hline
$\textit{walking\_xyz} $       & 0.7521   & 0.0932   & 0.015     & 0.0247   & \textbf{0.014}   \\
\hline
$\textit{walking\_rpy}$        & 0.8705   & 0.2333   & \textbf{0.035}  & 0.4442   & 0.0871   \\
\hline
$\textit{walking\_halfsphere}$ & 0.4863   & 0.0811    & \textbf{0.025}  & 0.0303   & 0.0307   \\
\hline
\hline
Method & Conventional   & Geometric  & Semantic + geometric & Semantic + geometric & Semantic + optical flow \\    
\hline
\hline
\end{tabular}}
\end{table*}

\begin{table*}[th!]
\caption{RTE for TUM dataset}
\label{Table 2}
  \centering
  \resizebox{\linewidth}{!}{ 
\renewcommand{\arraystretch}{2.0}
\begin{tabular}{M{0.2\linewidth}|M{0.08\linewidth} M{0.08\linewidth} M{0.08\linewidth} M{0.08\linewidth} M{0.08\linewidth} M{0.08\linewidth} M{0.08\linewidth} M{0.08\linewidth} M{0.08\linewidth} M{0.08\linewidth}}
\hline
\hline
\multirow{2}{*}{\textbf{Sequence}} & \multicolumn{2}{c}{\textbf{ORB-SLAM2}} & \multicolumn{2}{c}{\textbf{SLAM with motion removal}} &  \multicolumn{2}{c}{\textbf{Dyna-SLAM}} & \multicolumn{2}{c}{\textbf{DS-SLAM}} & \multicolumn{2}{c}{\textbf{Our method}}  \\ \cline{2-11}`
                                & RMSE (m) & RMSE (deg) & RMSE (m) & RMSE (deg) & RMSE (m) & RMSE (deg) & RMSE (m) & RMSE (deg) & RMSE (m) & RMSE (deg)  \\ 
\hline
$\textit{walking\_xyz}$         & 0.4124 & 7.7432 & 0.1214 & 3.2346 & - & - & 0.0333 & 0.8266 & \textbf{0.0182}   & \textbf{0.5942} \\
\hline
$\textit{walking\_rpy}$         & 0.4249 & 8.0802 & 0.3077 & 4.4968 & - & - & 0.1503 & 3.0042 & \textbf{0.0779} & \textbf{1.9477}  \\
\hline
$\textit{walking\_halfsphere}$  & 0.3550 & 7.3744 & 0.0955 & 3.3035 & - & - & 0.0297 & 0.8142 & \textbf{0.0183} & \textbf{0.4205}  \\
\hline
\hline
Method                    & \multicolumn{2}{c}{Conventional}   & \multicolumn{2}{c}{Geometric}  & \multicolumn{2}{c}{Semantic + geometric} & \multicolumn{2}{c}{Semantic + geometric} & \multicolumn{2}{c}{Semantic + optical flow} \\                
\hline
\hline
\end{tabular}}
\end{table*}

\subsection{TUM dataset}
The TUM RGB-D dataset offers multiple sequences of images with ground truth trajectories in an indoor environment, comprising RGB images and depth maps. It is widely used as a benchmark for evaluating dynamic VSLAM. Since most dynamic objects in TUM are defined in the training dataset for detection tasks, supervised models can effectively detect them. Consequently, many dynamic VSLAM algorithms employing segmentation models perform well on the TUM dataset. Three sequences are used to compare localization performance: 
1) a half-sphere trajectory ($\textit{walking\_halfsphere}$), 2) translational ($\textit{walking\_xyz}$), and 3) rotational ($\textit{walking\_rpy}$) movements. 

Tables \ref{Table 1} and \ref{Table 2} show the ATE and RTE comparison between conventional SLAM, non-semantic dynamic SLAM with geometric constraint, semantic-based dynamic SLAM with a supervised model, and the proposed SLAM. For RTE, the time step interval $\Delta$ is set to equal the frequency of the camera frame, so that the RTE compares pose change between one-second intervals. Recall that the performance of dynamic segmentations extraction is depicted in Fig. \ref{Dynamic Mask}. 

Although ORB-SLAM2 employs a robust estimation algorithm to exclude a small number of outliers, it lacks a feature classification algorithm, resulting in higher errors compared to dynamic SLAM methods in environments with significant dynamic factors. VSLAM with motion removal \cite{SUN2017110} uses image differencing and particle filter to exclude dynamic areas. Even though non-semantic dynamic VSLAM with geometric constraint gets better performance than conventional VSLAM and does not require semantic prior knowledge, its error is greater than semantic-based methods. Since the dynamic factors in these sequences are people, which are predefined as movable objects in the training dataset, both dynamic VSLAM with a supervised model versus our proposed method classify the static versus dynamic features similarly. They both exhibit better performance than ORB-SLAM2 as well as non-semantic dynamic VSLAM with a geometric constraint on the TUM dataset of a known class of moving objects.

For example, Table \ref{Table 1} shows that the ATE of ORB-SLAM2 in $\textit{walking\_xyz}$ sequence is greater than 0.7 meters, geometric-based VSLAM's error is less than 0.1, and the semantic dynamic VSLAM and proposed VSLAM have the ATE of less than 0.025 meters. In all sequences, the proposed VSLAM shows comparable performance with the existing dynamic VSLAM with a supervised model. 
The best performances are noted as bold numbers in the tables. 
From Table \ref{Table 2}, the proposed method has the lowest RTE error over other dynamic SLAM methods (Dyna-SLAM did not report RTE evaluation).

\begin{figure} [hb!]    
    \subfloat[ATE from ORB-SLAM2]{\includegraphics[width=0.33\textwidth]{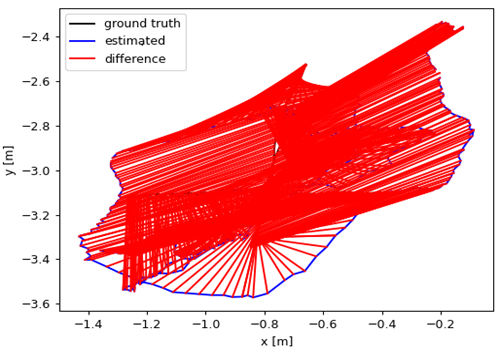}} %
    \subfloat[ATE from DS-SLAM]{\includegraphics[width=0.33\textwidth]{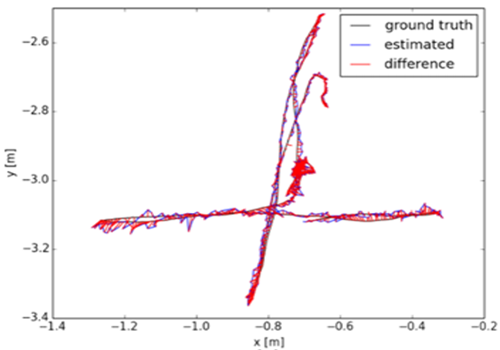}} %
    \subfloat[ATE from our method]{\includegraphics[width=0.33\textwidth]{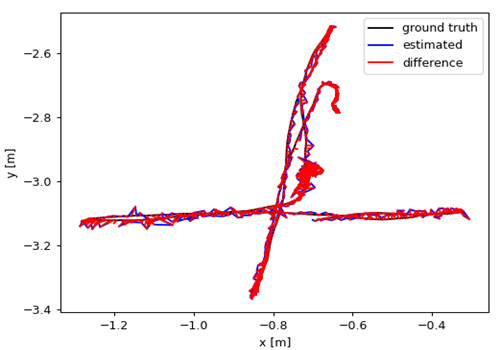}} %
    \hfill         
    \subfloat[RTE from ORB-SLAM2]{\includegraphics[width=0.33\textwidth]{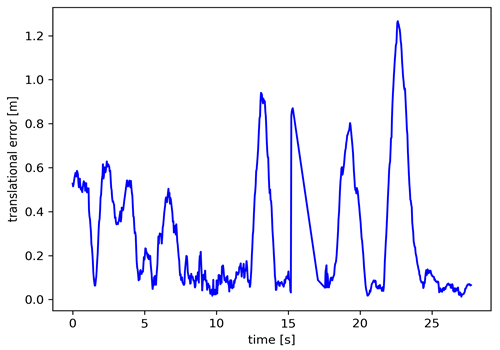}} %
    \subfloat[RTE from DS-SLAM]{\includegraphics[width=0.33\textwidth]{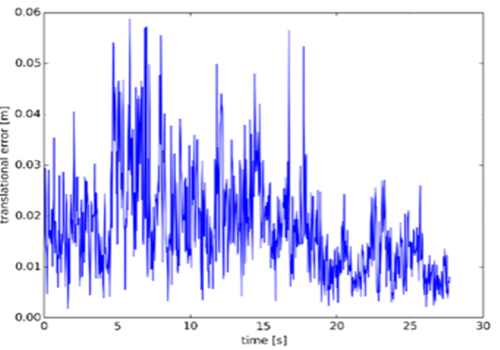}} %
    \subfloat[RTE from our method]{\includegraphics[width=0.33\textwidth]{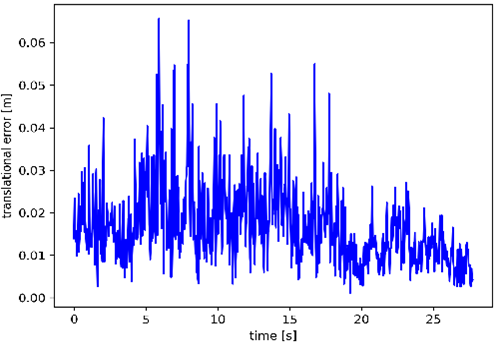}} %
\caption{ATE and RTE for TUM RGB-D dataset}
\label{ATE and RTE for TUM}
\end{figure}

Fig. \ref{ATE and RTE for TUM} visualizes the absolute and relative accuracy of position estimation for the $\textit{walking\_xyz}$ sequence. Because the $\textit{walking\_xyz}$ sequence has a small z-direction translation, only xy translation is drawn in the figure. From the figure, ORB-SLAM2 shows a large positional error in a dynamic environment because conventional SLAM does not have a feature classification algorithm. On the contrary, DS-SLAM and our method have a much smaller error. This improvement comes from the feature classification result that both dynamic SLAMs exclude features in movable objects for the camera pose estimation.

\subsection{OMD dataset}
The OMD dataset also provides multiple sequences of images in an indoor environment, comprising stereo RGB images and depth maps. The difference with TUM dataset is that the OMD dataset involves objects that are not classified a priori. Specifically, OMD includes dynamic swing boxes, that are not labeled in the detector dataset, and so the supervised models fail to detect those. Two sequences are used to compare localization performance: Four boxes swinging in a room with the camera having different movements: 1) translational (\textit{Translational}), and 2) unconstrained (\textit{Unconstrained}) movements. 

\begin{table}[ht!]
\caption{ATE for OMD dataset}
\label{Table 3}
  \centering
  \resizebox{\linewidth}{!}{ 
\renewcommand{\arraystretch}{2.0}
\begin{tabular}{M{0.2\linewidth}|M{0.4\linewidth} M{0.4\linewidth}}
\hline
\hline
\multirow{2}{*}{\textbf{Sequence}} & \textbf{ORB-SLAM2} & \textbf{Our method} \\ \cline{2-3}`
                           & RMSE (m)  & RMSE (m)    \\ 
\hline
\textit{Translational}     & 0.0560 & 0.0355  \\
\hline
\textit{Unconstrained}     & 0.0603 & 0.0356 \\
\hline
\hline
\end{tabular}}
\end{table}

\begin{figure} [ht!]
\centerline{\includegraphics[width=0.45\textwidth]{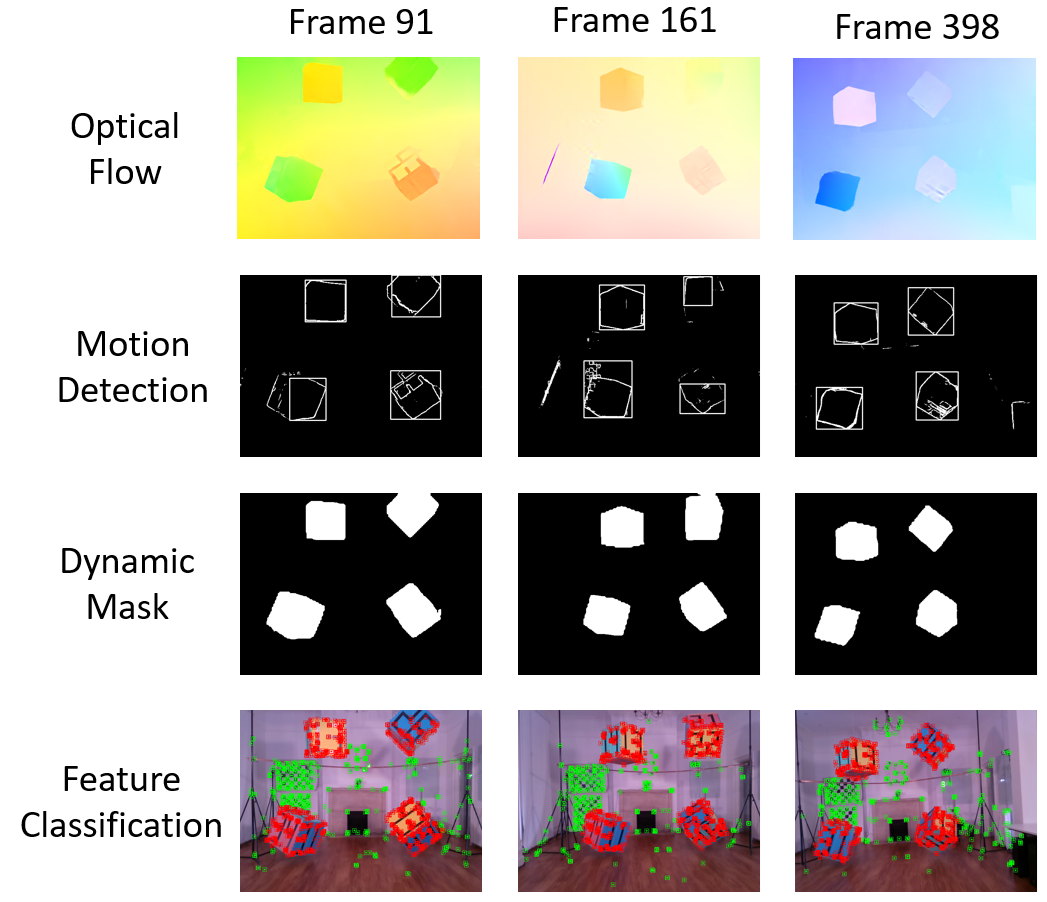}}
\caption{Dynamic mask and feature classification result}
\label{fig: Dynamic Mask From Flow}
\end{figure}

Most dynamic VSLAM with a supervised model do not report their performance for unknown objects quantitatively. We evaluate the proposed method and compared its performance to the conventional VSLAM, ORB-SLAM2. Table \ref{Table 3} shows that our method obtains better localization performance by improving feature classification. This improvement comes from the dynamic features obtained from a high optical flow gradient as introduced by us. Because the swing boxes are not included in the training dataset, the supervised models are not able to detect them. However, the swing boxes get detected from our proposed high optical flow gradient. Recall that the performance of static versus dynamic feature identification is shown in Fig. \ref{Dynamic Mask}, where the swing boxes are not detected under a supervised segmentation. However, the proposed motion boundaries from high optical flow gradients are able to detect the swing boxes, showing the contribution of the proposed method. Fig. \ref{fig: Dynamic Mask From Flow} shows more examples of feature classification in \textit{Translation} sequence. The dynamic features identified from optical flow gradient are more precise compared to any other existing approach in the literature. 

\section{Conclusion}
This paper has introduced a semantic VSLAM system capable of detecting static versus dynamic features from images containing both known and unknown moving objects. The proposed method yields accurate static vs. dynamic feature classification and camera motion tracking in dynamic environments containing unknown object types. The main novel contribution is the use of optical flow gradient information together with unsupervised segmentations and supervised object detection, coupled with iterative consistency check for outlier identification and exclusion. Comparative evaluations against the existing dynamic semantic VSLAM methods indicate that the proposed method offers improvement in accuracy for images with also unknown moving objects, whereas it offers comparable accuracy for images with known moving objects. Given the diverse array of objects encountered in real-world scenarios, reducing reliance on predefined classes is crucial that our approach facilitates.

Future extensions of this work may involve enhanced developing path-planning capabilities in dynamic environments based on the proposed VSLAM. Also the approach utilized by 
\cite{RDMO-SLAM}, which employs a semantic thread for keyframes and predicts dynamic segmentations using optical flow for other keyframes, could be further refined using our approach to enhance the accuracy and real-time performance.

\bibliographystyle{unsrtnat}
\bibliography{reference}

\end{document}